\documentclass{article}

\usepackage{spconf,amsmath,graphicx}
\usepackage{setspace}
\usepackage{amssymb}
\usepackage{subcaption} 
\usepackage{caption}

\title{Enhancing Generative Class Incremental Learning Performance with Model Forgetting Approach}
\name{Taro Togo$^{\dag}$, Ren Togo$^{\dag\dag}$, Keisuke Maeda$^{\dag\dag\dag}$, Takahiro Ogawa$^{\dag\dag}$ \ and Miki Haseyama$^{\dag\dag}$}
\address{$^{\dag}$Graduate School of Information Science and Technology, Hokkaido University, Japan\\
$^{\dag\dag}$ Faculty of Information Science and Technology, Hokkaido University, Japan\\
$^{\dag\dag\dag}$ Data-Driven Interdisciplinary Research Emergence Department, Hokkaido University, Japan\\
E-mail: \{taro\_togo, togo, maeda, ogawa, mhaseyama\}@lmd.ist.hokudai.ac.jp
}

\begin{document}
\ninept
\maketitle
\begin{abstract}
This study presents a novel approach to Generative Class Incremental Learning (GCIL) by introducing the forgetting mechanism, aimed at dynamically managing class information for better adaptation to streaming data. GCIL is one of the hot topics in the field of computer vision, and this is considered one of the crucial tasks in society, specifically the continual learning of generative models.
The ability to forget is a crucial brain function that facilitates continual learning by selectively discarding less relevant information for humans. However, in the field of machine learning models, the concept of intentionally forgetting has not been extensively investigated. In this study we aim to bridge this gap by incorporating the forgetting mechanisms into GCIL, thereby examining their impact on the models' ability to learn in continual learning. Through our experiments, we have found that integrating the forgetting mechanisms significantly enhances the models' performance in acquiring new knowledge, underscoring the positive role that strategic forgetting plays in the process of continual learning.

\end{abstract}
\begin{keywords}
Continual learning, class incremental learning, machine unlearning, generative models.
\end{keywords}
\section{Introduction}
\label{sec:intro}
The rapid advancements in deep learning have led to a performance in image recognition and natural language processing that equals or surpasses human capabilities in certain specific tasks~\cite{chang2023survey}. In recent years, deep learning models are beginning to be employed to address various real-world challenges. Particularly, image generation models, benefiting from learning on large-scale data and novel architectures such as Transformers~\cite{vaswani2017attention} and Diffusion models~\cite{ho2020denoising}, are anticipated to have a wide range of applications. As efforts to implement deep learning in practical applications for a long time, continual learning has begun to gain attention~\cite{de2021continual}. Real-world data is not static but often in a stream format, where new types of data are constantly generated. Learning techniques for stream data has become one of the important challenges in continual learning.
Continual learning refers to the approach of acquiring new knowledge, such as adaptation to new domains or learning new class identifications, while retaining the knowledge already learned. This ensures the long-term functionality of the constructed models.

Class incremental learning (CIL) is one of the tasks in continual learning and a learning approach that sequentially recognizes new classes in stream data~\cite{masana2022class}. In CIL, the suppression of a phenomenon known as catastrophic forgetting becomes a crucial challenge~\cite{kirkpatrick2017overcoming}. When learning new classes, direct training of the existing network can lead to the forgetting of knowledge about previously learned classes, resulting in irreversible degradation of the model's performance. Therefore, when constructing CIL models, the critical issue is how to retain previous knowledge and suppress catastrophic forgetting while accommodating new classes.

Within the framework of continual learning~\cite{van2021class,liu2020generative}, the primary focus of this study is exploring Generative Class Incremental Learning (GCIL).
This approach aims to learn the generation of images for new classes while preserving the quality of generation for existing classes.
GCIL is an emerging task that has risen in prominence alongside advancements in image generation technology. It represents one of the key tasks for the continuous learning of generative models. On the other hand, GCIL also faces the same challenge of catastrophic forgetting as conventional CIL. It is necessary to establish a new learning strategy that takes into account catastrophic forgetting in the generation task.

In GCIL, previous approaches have focused on refining the quality of generated images. These approaches often involve adding constraint terms to the existing model parameters or using images generated from the previous generative models when learning new classes~\cite{guan2019dcigan,li2020incremental}. 
Specifically, in the field of image generation, various techniques have been proposed to enable the generation of desired images using class information, segmentation, or other guides~\cite{li2021semantic,li2022union}. These techniques have found success in conditional image generation and personalization by leveraging existing knowledge~\cite{ruiz2023dreambooth}. Although the primary focus of this research on image generation models has been to enhance the quality of the images they generate, there remains considerable potential for further investigation into the continuous training of these models and the mitigation of secondary effects that emerge throughout this process.

To examine secondary effects that may arise during the learning process of GCIL, we focus on generative models with a limited number of parameters. In this scenario, the generative models are required to effectively learn and generate representations for new classes without affecting the existing ones. Furthermore, it is critical to ensure that the deletion or modification of representations for any existing class minimizes the influence on the remaining classes and does not restrain the model's ability to learn new classes effectively. 

Recently, a method of Selective Amnesia (SA) has been proposed in the field of image generation, focusing on the concept of forgetting~\cite{heng2023selective}. SA ingeniously utilizes the learning process of continual learning to successfully make the model forget specific concepts. This approach has been proposed for replacing images related to copyrighted or harmful content in image generation models, even in the absence of original training data. Heng et al.~\cite{heng2023selective} argue that this approach allows for the provision of a safer and more ethical output of the image generation model. 
On the other hand, SA could be viewed as a mechanism for resetting the knowledge related to existing classes in image generation models.
They also reported that the phenomenon known as ``Concept Leakage" was observed, where forgetting one concept inadvertently affects similar or encompassing concepts. Thereby, this observation suggests that the knowledge related to existing classes in the models may influence the learning of new classes. 
Actually, this approach is very similar to the human learning mechanism~\cite{Niv2019-ed}. The human brain, with its physical capacity limitations, requires the process of forgetting old information to acquire new knowledge. 
Inspired by this human learning mechanism, we assert the significance of forgetting mechanisms in machine learning models. 

\begin{figure}[t]
\centering
\includegraphics[width=0.48\textwidth]{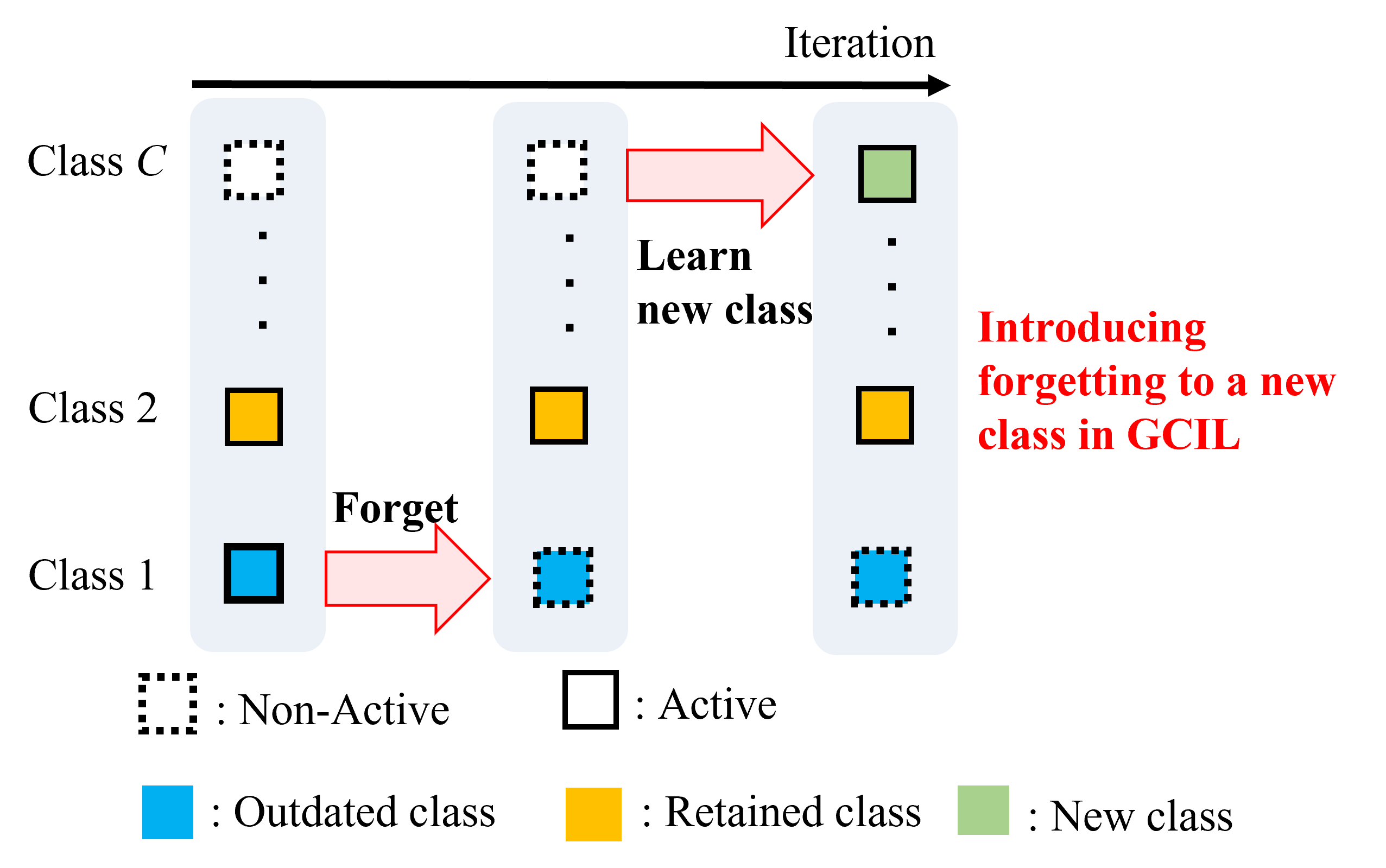} 
\caption{Overall architecture and our problem setting of generative class incremental learning with forgetting.}
\label{fig:schematic-diagram}
\end{figure}

In this paper, we propose a novel method by introducing the forgetting mechanisms to enhance the model performance in GCIL (Fig.~\ref{fig:schematic-diagram}). 
Specifically, we examine the impact of forgetting mechanisms and concept leakage phenomenon on the performance of models when learning new data with a fixed capacity. 
Our proposed method suggests new possibilities in GCIL and aims to contribute to the long-term performance improvement of these models.
Building on this conceptual foundation, we introduce a novel GCIL method with a forgetting mechanism. The proposed method employs a selective forgetting process in pre-trained image generation models to reset and manage classes. The reset mechanism focuses on removing outdated or irrelevant classes from the model's memory, promoting the learning of new classes. By retraining models with this framework, we improve their efficiency in learning and retaining new classes, enhancing adaptability in evolving data environments. Our method promotes the development of flexible image generation models, capable of dynamically updating class information while efficiently discarding non-essential classes. This balance of forgetting and learning is crucial in the rapidly changing field of image generation, meeting the demands of data-driven applications.

The key contributions of our method are shown as follows.
\begin{enumerate}
\item \textbf{Introducing forgetting mechanism in GCIL:}
We have pioneered the incorporation of the forgetting mechanism into GCIL. This novel integration allows for more dynamic management of class information within image generation models, tailoring them to adapt to streaming data more effectively.

\item \textbf{Enhancing GCIL performance by forgetting:} We provide empirical evidence that the introduction of the forgetting mechanism can lead to significant improvements in GCIL performance. This finding underscores the utility of strategically removing outdated or irrelevant class information to enhance overall learning performance.

\item \textbf{Elucidating forgetting potential in diverse scenarios:} We extend the concept of forgetting beyond traditional applications, demonstrating its potential and versatility in a range of diverse scenarios. This exploration opens up new avenues for applying forgetting techniques in various scenarios, thereby broadening the applicability of GCIL methods.
\end{enumerate}

\section{Related Work}
\label{sec:related}
In this section, we provide an overview of existing research related to continual learning, CIL, and machine unlearning. We also explain how these fields are interrelated, and identify their relevance to our approach to address the challenges of learning and forgetting in generative models.

\subsection{Continual Learning}
Continual learning indicates a model's ability to progressively learn from and adapt to new tasks and data over time, while also efficiently managing its knowledge to retain relevance and functionality~\cite{de2021continual,hadsell2020embracing}. This concept is particularly vital in dynamic environments where data types evolve, and new categories emerge. The essence of continual learning lies in its ability to make models robust and adaptable to the ever-changing landscape of real-world scenarios where static training methods prove insufficient.

In this broader context, continual learning involves not just the adaptation to new data, but also the strategic removal of obsolete or irrelevant information~\cite{aljundi2019task}. This approach is essential for maintaining an up-to-date and efficient model. To tackle these challenges, various strategies are employed, including model manipulation and reorganization based on data replacement~\cite{felps2020class, JMLR:v21:19-253}. These strategies enhance the model's ability to stay relevant in the face of changing data landscapes while ensuring efficient resource utilization.

Nevertheless, continual learning faces significant challenges. Catastrophic forgetting remains a major hurdle, where models lose previously acquired knowledge upon learning new information~\cite{hu2019overcoming}. Additionally, the field faces the stability-plasticity dilemma, which involves striking a balance between the absorption of new knowledge and the retention of existing knowledge. These challenges underscore the complexity of developing models capable of continual learning, emphasizing the need for innovative solutions that address both the incorporation of new information and the judicious pruning of outdated data.

\subsection{Class Incremental Learning}
CIL represents a crucial approach in the field of machine learning, focusing on the sequential introduction and integration of new classes into an existing model~\cite{masana2022class}. This method is vital for empowering models to adapt and evolve in response to emerging information. Within the spectrum of CIL techniques, Elastic Weight Consolidation (EWC)~\cite{kirkpatrick2017overcoming} and Gradient Episodic Memory (GEM)~\cite{lopez2017gradient} stand out as key methods. These strategies enhance the model's flexibility, by concentrating on the retention of pre-existing knowledge while simultaneously assimilating new classes. Such an approach ensures the model's robustness in various learning environments and the dynamic nature of real-world scenarios. A primary challenge in CIL is mitigating catastrophic forgetting, a common issue when new classes are incorporated~\cite{song2024overcoming}. This challenge requires a careful balance between integrating new information and preserving existing knowledge to ensure that the model retains previously acquired skills and data during expansion and adaptation.

With the rapid advancement of generative models in recent years, image-generation models are becoming widely prevalent in society. The foundation of these models lies in large-scale data and sophisticated modeling techniques. However, these models face challenges in continuous learning and usage, necessitating a collaborative approach with continual learning methodologies. Therefore, to effectively and continuously utilize machine learning models that are increasingly permeating society, exploring tasks like Generative Class Incremental Learning (GCIL) becomes more critical~\cite{van2021class,liu2020generative}.

\subsection{Machine Unlearning}
Machine unlearning is a field within machine learning designed to selectively eliminate specific data, such as copyrighted or harmful images, from existing models, while effectively addressing privacy and ethical concerns~\cite{xu2023machine}. A major challenge in this process is the cost of relearning, often necessitating access to the original dataset~\cite{cao2015towards}. As learning data becomes more extensive, the task of data deletion grows increasingly challenging. Thus, machine unlearning has primarily focused on classification problems. However, this study extends its application to generative tasks, allowing for the conceptualization of machine unlearning in the context of generative models.

Advancements in image generation techniques like Selective Amnesia (SA)~\cite{heng2023selective} have facilitated the forgetting or replacing data without the original dataset, relying solely on retraining. This evolution represents a significant advancement in continual learning, providing a more efficient and feasible means to manage and update knowledge embedded within models. It is noteworthy that SA primarily focuses on the forgetting of existing classes and does not inherently account for the potential impact on other classes.

\section{Generative Class Incremental Learning with Forgetting}

\begin{figure*}[t]
\centering
\includegraphics[width=1\linewidth]{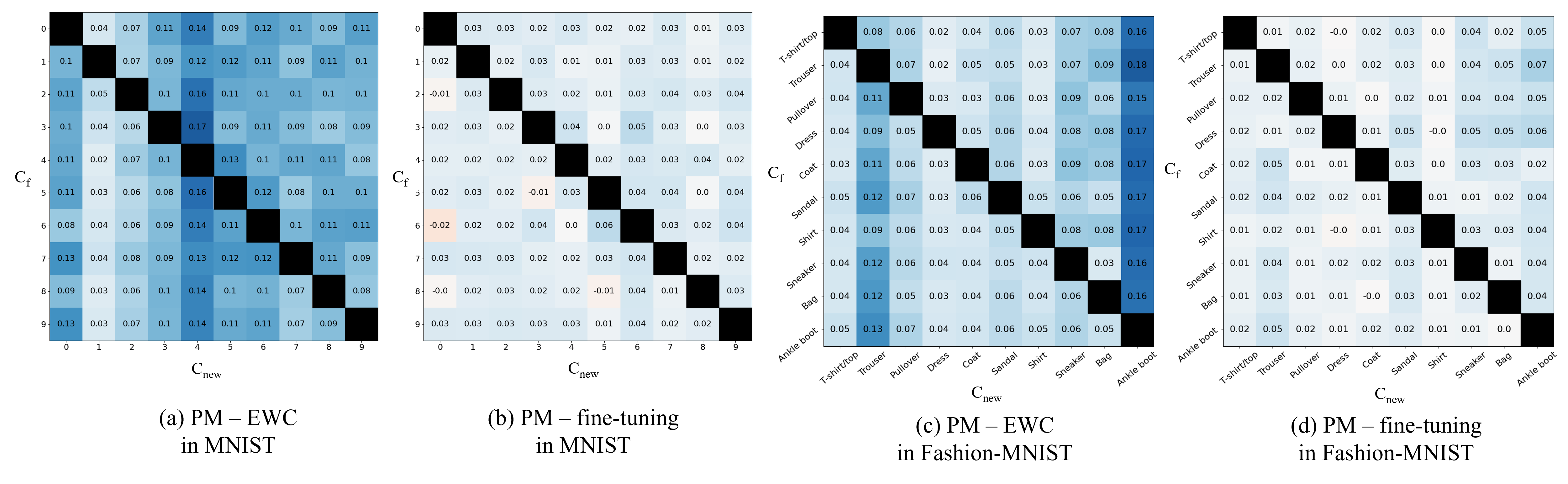}

\caption{The changes in the probability of the class are influenced by the integration of the PM (white noise) into different learning techniques when new classes are learned in various datasets. (a) and (b) show the results for the MNIST dataset, while (c) and (d) show the results for the Fashion-MNIST dataset. The vertical axis indicates the classes subject to forgetting, whereas the horizontal axis represents the newly acquired classes. Blue, red, and black colors signify positive impact, negative impact, and absence on the data, respectively.}
\label{fig:conf_matrix_comparison}
\end{figure*}
\subsection{Overall Architecture}
In this section, we introduce a novel continual learning scheme for a deep generative model. This scheme consists of two steps ``Forgetting", and ``Learning". Our approach focuses on utilizing a specific image generation model to achieve the following three specific objectives.

First, we define a generative model \( M \)  capable of generating images of classes \(c=1,2, \cdots, C\). The model \( M \) is designed to generate images corresponding to class information, undergoing a forgetting process to inhibit the generation of images for a specific class. The class to be forgotten is denoted as \( c_{\rm f} \), and the model after the forgetting process for class \( c_{\rm f} \) is denoted as \( M_{\rm f} \).
The model \( M_{\rm f} \) is a model that loses the ability to generate images for class \( c_{\rm f} \), while retaining the capability to generate images for the existing classes \( c_{\rm r} \).

Subsequently, for the post-forgetting model \( M_{\rm f} \), we train the model \( M_{\rm final} \) to incorporate a new class \( c_{\rm new} \) as GCIL. \( M_{\rm final} \) is trained to acquire the capability to generate images for the newly introduced class \( c_{\rm new} \). 
The outlined workflow is derived from the dataset \( D \) as described below.

\begin{equation}
\begin{split}
  D & = D_{\rm f} \cup D_{\rm r} \cup D_{\rm new} \\
    & = \{ (x^{(n)}_{\rm f}, c^{(n)}_{\rm f}) \ | \ n = 1, \ldots, N_{\rm f} \} \\
    & \quad \cup \{ (x^{(m)}_{\rm r}, c^{(m)}_{\rm r}) \ | \ m = 1, \ldots, N_{\rm r} \} \\
    & \quad \cup \{ (x^{(l)}_{\rm new}, c^{(l)}_{\rm new}) \ | \ l = 1, \ldots, N_{\rm new} \},
\end{split}
\label{eq:dataset}
\end{equation}
where  \( x \) indicate the image and \( N_{\rm f} \), \( N_{\rm r} \), and \( N_{\rm new} \) indicate the number of images in each subset.
In the proposed method, we construct the forgetting model \( M_{\rm f} \) by applying the improved SA model. SA allows the embedding of meaningless image information, such as white noise images, into the post-learning model for intended classes, thereby facilitating the forgetting of generative information related to specific classes. Our approach elucidates that in the context of CIL, enabling the pre-trained model to forget information simplifies the learning of new classes \( c_{\rm new} \). During the steps of GCIL for the training of \( M_{\rm final} \), we employ EWC, one of the most popular approaches in continual learning. EWC efficiently learns and integrates new classes into the model without significantly disrupting the existing knowledge structure.

\subsection{Introducing Forgetting Mechanism}
In the proposed method, we introduce the forgetting mechanism into GCIL through SA for managing the dataset \(D\), with a particular focus on the class information of \(D_{\rm f}\) to build the forgetting model \( M_{\rm f} \). SA introduces a novel paradigm in image generation models by enabling the selective forgetting of specific data. This technique facilitates the efficient removal of information corresponding to the forgetting model \( M_{\rm f} \).

Specifically, SA employs Generative Replay~\cite{shin2017continual} to ensure that information fitting to classes that should be remembered is targeted to images generated by previous iterations of the generator. Meanwhile, outdated information is targeted for learning with specific images marked for forgetting. This process allows for a dynamic adjustment of the model, taking into account the significance of each class's information to prevent the improper forgetting of previous classes during forgetting phases.

In the proposed method, SA is utilized not merely as a tool for erasing outdated classes \( c_{\rm f} \) but as a means to transform these classes into entities that enhance the model's effectiveness in learning unknown new classes. This replacement strategy contributes to a more refined and targeted learning process for new classes \( c_{\rm new} \).
First, the Bayesian probability formulation for updating the parameters \(\theta\) of a model \(M\) after learning from datasets \(D_{\rm f}\) and \(D_{\rm r}\) is defined as
\begin{align}
    \log p(\theta | D_{\rm f}, D_{\rm r}) &= \log p(D_{\rm f} | \theta, D_{\rm r}) + \log p(\theta | D_{\rm r}) - \log p(D_{\rm f} | D_{\rm r}) \nonumber \\
    &= \log p(D_{\rm f} | \theta) + \log p(\theta | D_{\rm r}) + \rm{Const}.
\end{align}
The constant term \rm{Const} accounts for the normalization factor that arises from the conditional probability of data \(D_{\rm f}\) and \(D_{\rm r}\). Then the posterior conditioned on only \(D_{\rm r}\) can be expressed as:
\begin{align}
    \log p(\theta | D_{\rm r}) &= \log p(D_{\rm f} | \theta) + \log p(\theta |D_{\rm f}, D_{\rm r}) + \rm{Const} \nonumber\\
    &= \log p(x_{\rm f} | \theta, c_{\rm f}) -  \lambda \sum_i \frac{F_i}{2} (\theta_i - \theta_i^*)^2 + \rm{Const},
    \label{eq:forgetting}
\end{align}
where \( F \) represents the Fisher Information Matrix (FIM) and \(i \in \{\mathrm{f}, \mathrm{r}\}\)
and  \( \lambda \) serves as a tuning parameter that adjusts the penalty's strength. Since the goal is to update the parameters of the pre-trained model, it is impossible to sample directly from \(D_{\rm r}\) to optimize Eq.~(\ref{eq:forgetting}) via the Evidence Lower Bound (ELBO) optimization. Instead, we achieve forgetting the class \(c_{\rm f}\) through pseudonymous sampling via Generative Replay~\cite{shin2017continual}. For more details,  refer to the literature~\cite{heng2023selective}. Through this process, the construction of forgetting model \( M_{\rm f} \) can be achieved.

\subsection{Generative Class Incremental Learning}
In the context of our GCIL method, particularly after the application of forgetting mechanisms for \(M_{\rm f}\), we adopt EWC as a strategic approach to balance the acquisition of new class knowledge with the preservation of existing class information for the training of \( M_{\rm final} \). EWC employs a Bayesian framework to approximate the weight distribution,  thereby enabling the seamless assimilation of new classes denoted by \( D_{\rm new} \), while safeguarding the knowledge of previously learned classes represented by \( D_{\rm r} \), through the utilization of a pre-trained model, specifically, the forgetting model \(M_{\rm f}\) with \( \theta^* \).

EWC employs the Laplace approximation on the posterior distribution concerning \( D_{\rm r} \) toward the model \(M_{\rm f}\) parameterized by \(\theta^*\), which introduces a quadratic penalty to regulate the update rates of weights significant to \( D_{\rm r} \). This penalty mechanism is  formulated as like Eq.~(\ref{eq:forgetting}):

\begin{equation}
    \log p(\theta^* | D_{\rm r}, D_{\rm new}) = \log p(D_{\rm new} | \theta^*) - \lambda \sum_i \frac{F_i}{2} (\theta_i^* - \hat{\theta}_i)^2,
\end{equation}
where \( F \) represents the FIM, and \( \lambda \) serves as a tuning parameter that adjusts the penalty's strength.
To achieve computational efficiency, EWC adopts a diagonal approximation of the FIM, denoted by \( F_i \), which is defined as:

\begin{equation}
    F_i = \mathbb{E}_{p(D^*|\hat{\theta})} \left[ \left( \frac{\partial}{\partial \theta_i^*} \log p(D^*|\theta^*) \right)^2 \right].
\end{equation}
This approximation captures the sensitivity of each weight \( \theta_i^* \) to the model's output concerning the classes \(c_{\rm r}\) in \( D_{\rm r} \). The term \(D^*\) represents the union of  \(D_{\rm r}\) and \(D_{\rm new} \) . In the framework of variational models, the definition of \( F_i \) is further adapted to assess its influence on the ELBO, and it is expressed as

\begin{equation}
    F_i = \mathbb{E}_{p(x|\hat{\theta}, c)p(c)} \left[ \left( \frac{\partial}{\partial \theta_i^*} {\rm ELBO}(x|\theta^*,c) \right)^2 \right].
\end{equation}
Note that \(x\) represents image data from \(D\). These formulations underscore the nuanced role of EWC in balancing the learning of new classes \(c_{\rm new}\) while preserving the integrity of the existing classes \(c_{\rm r}\), through a carefully calibrated penalty mechanism and efficient computational strategies.
Through the adoption of EWC, our GCIL approach strategically mitigates the risk of catastrophic forgetting and ensures that the model remains adept at incorporating new classes without compromising the integrity and relevance of the existing class knowledge.

\begin{figure*}[t]
\centering
\begin{subfigure}[t]{0.24\linewidth}
    \centering
    \includegraphics[width=1\linewidth]{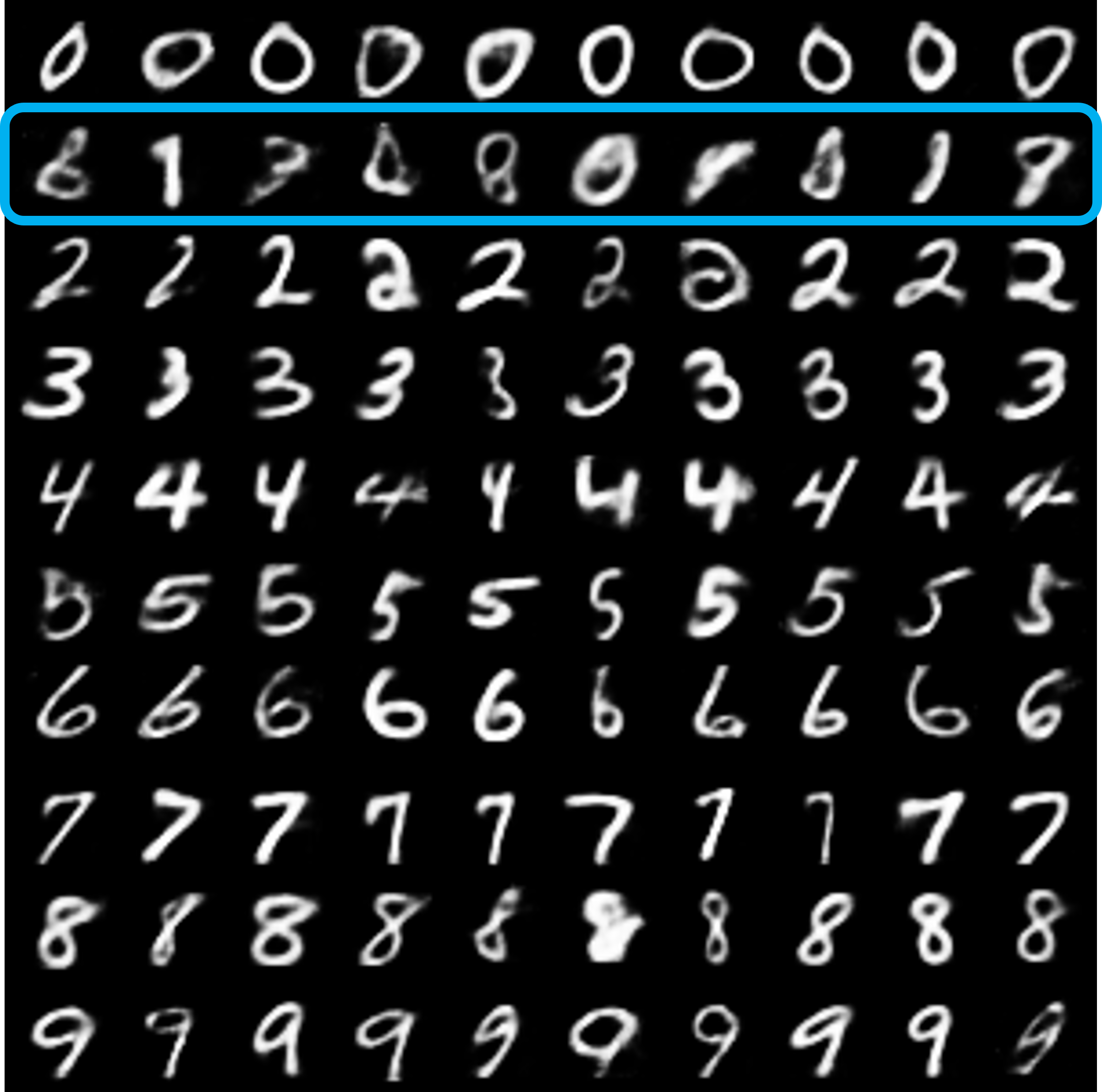}
    \caption{EWC}
    \label{fig:ewc_mnist}
\end{subfigure}
\hfill 
\begin{subfigure}[t]{0.24\linewidth}
    \centering
    \includegraphics[width=1\linewidth]{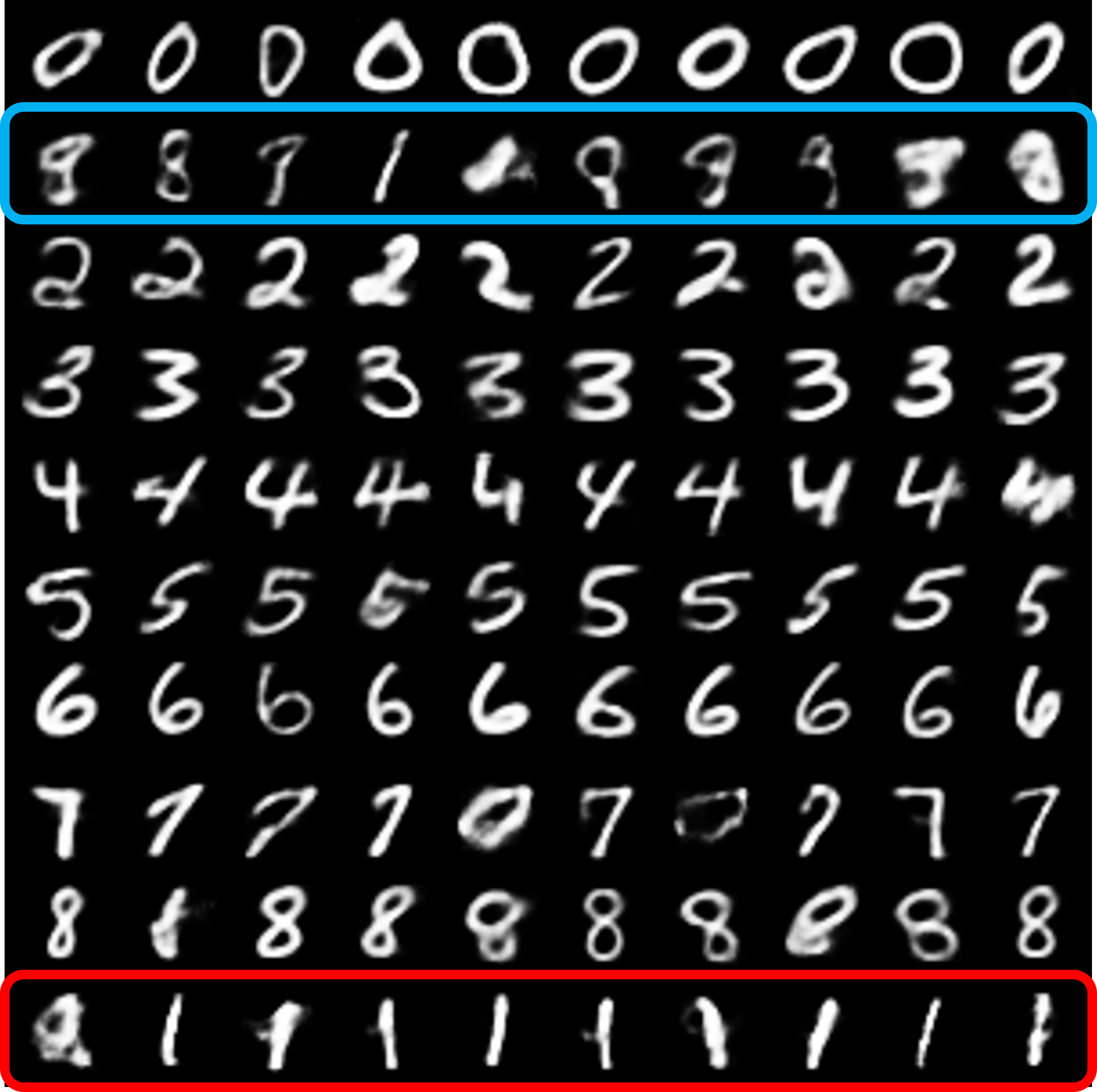}
    \caption{PM (embedding \(c_{\rm new}\))+EWC}
    \label{fig:sa_random_ewc_mnist}
\end{subfigure}
\hfill 
\begin{subfigure}[t]{0.24\linewidth}
    \centering
    \includegraphics[width=1\linewidth]{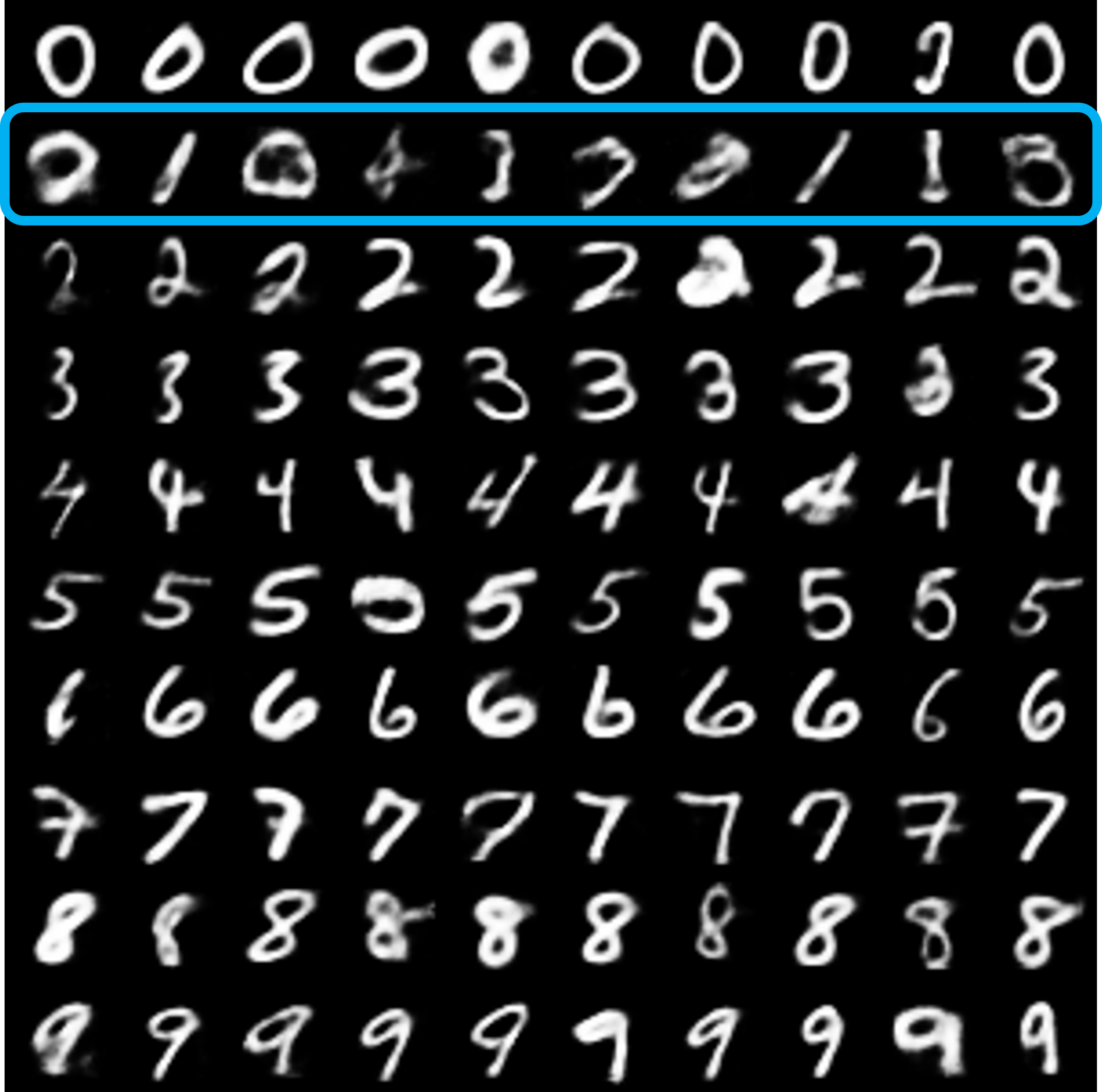}
    \caption{Fine-tuning}
    \label{fig:mnist_finetuning}
\end{subfigure}
\hfill
\begin{subfigure}[t]{0.24\linewidth}
    \centering
    \captionsetup{margin=-5pt}
    \includegraphics[width=1\linewidth]{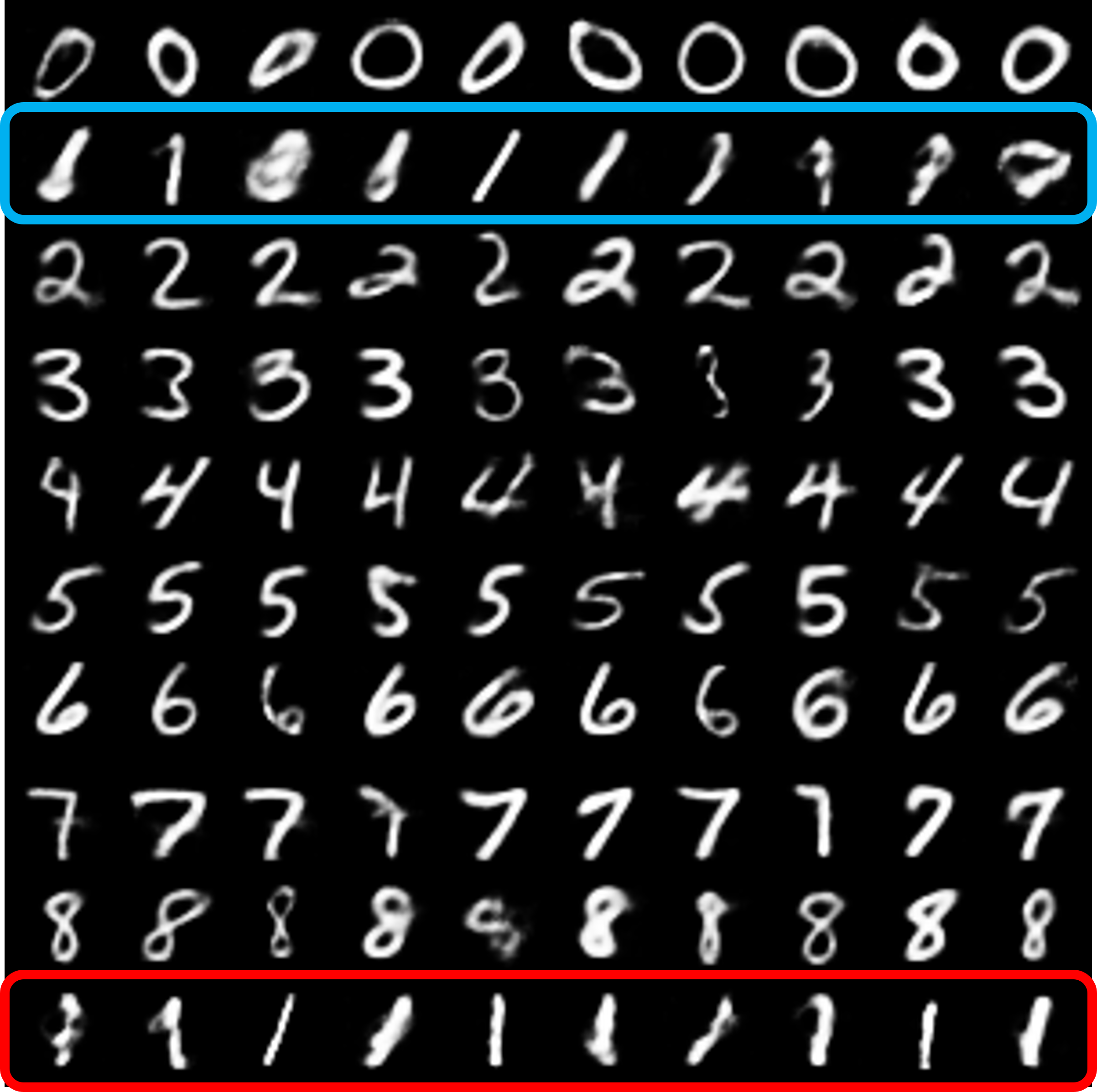}
    \caption{\hangindent=-20pt PM (embedding \(c_{\rm new}\))+fine-tuning}
    \label{fig:sa_white_noise_ewc_mnist}
\end{subfigure}
\hfill
\centering
\begin{subfigure}[b]{0.24\linewidth}
    \centering
    \includegraphics[width=1\linewidth]{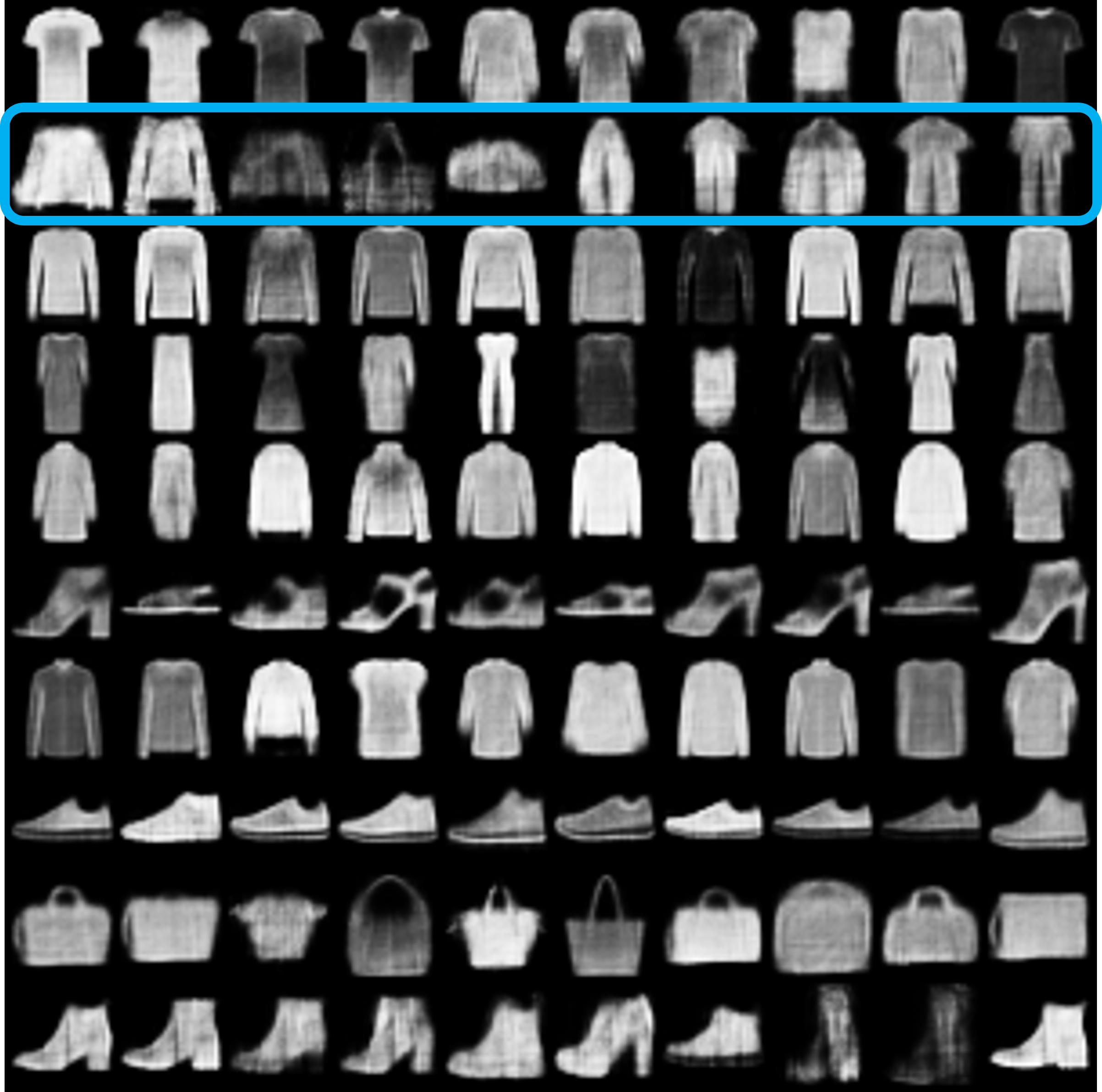}
    \caption{EWC}
    \label{fig:ewc_fashion}
\end{subfigure}
\hfill 
\begin{subfigure}[b]{0.24\linewidth}
    \centering
    \includegraphics[width=1\linewidth]{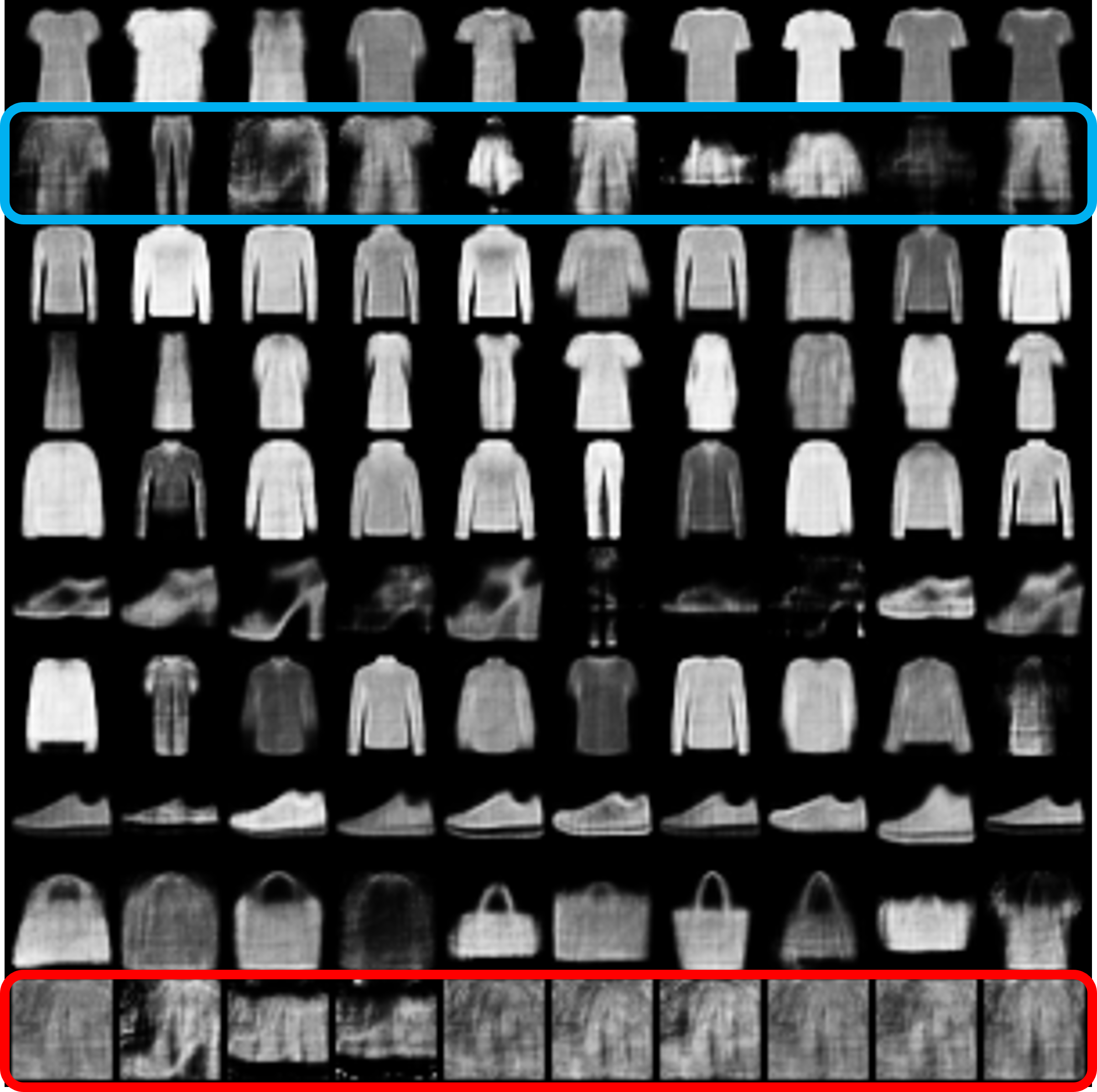}
    \caption{PM (white noise)+EWC}
    \label{fig:sa_random_ewc_fashion}
\end{subfigure}
\hfill 
\begin{subfigure}[b]{0.24\linewidth}
    \centering
    \includegraphics[width=1\linewidth]{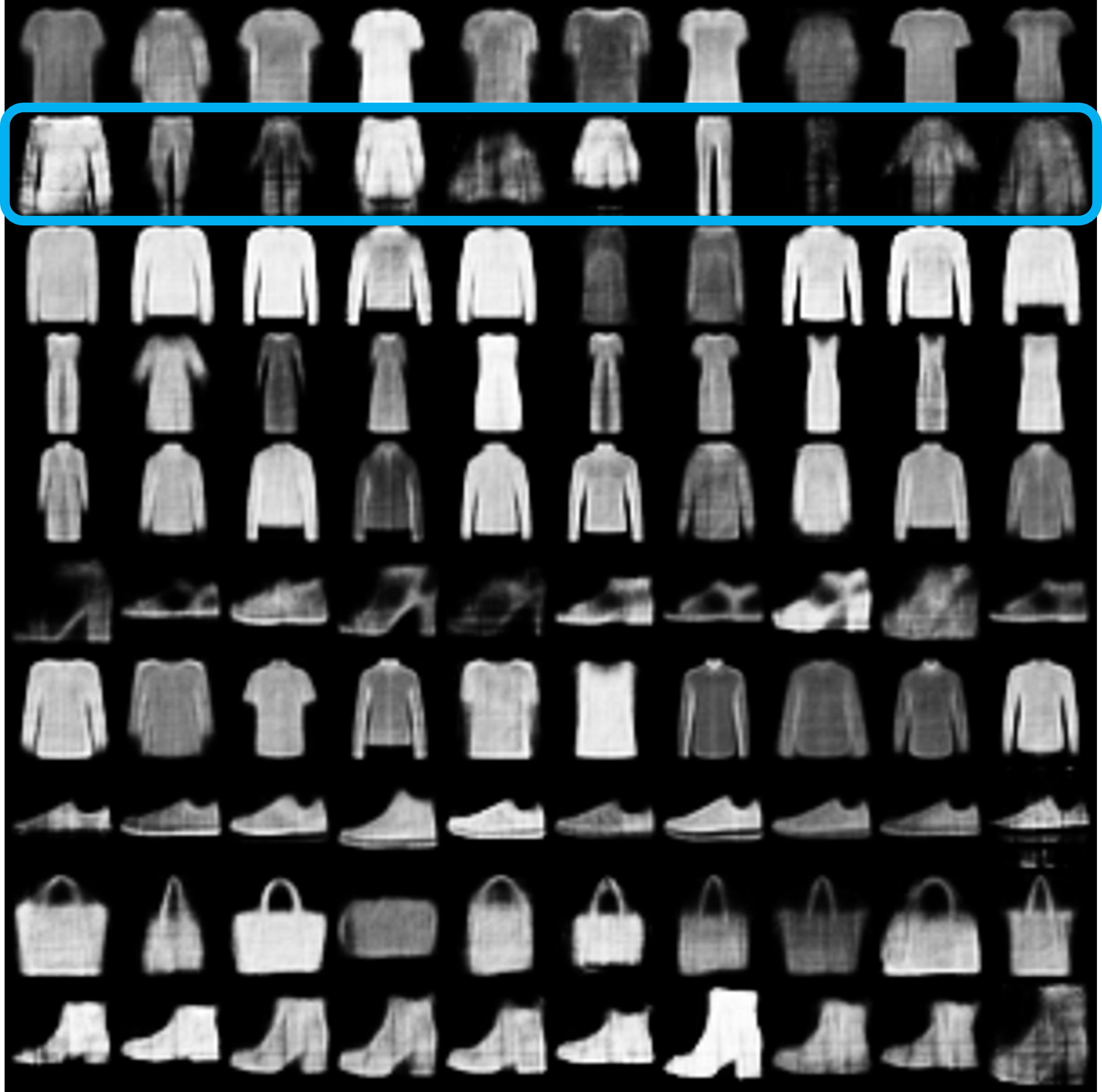}
    \caption{Fine-tuning}
    \label{fig:fashion_finetuning}
\end{subfigure}
\hfill
\begin{subfigure}[b]{0.24\linewidth}
    \centering
    \includegraphics[width=1\linewidth]{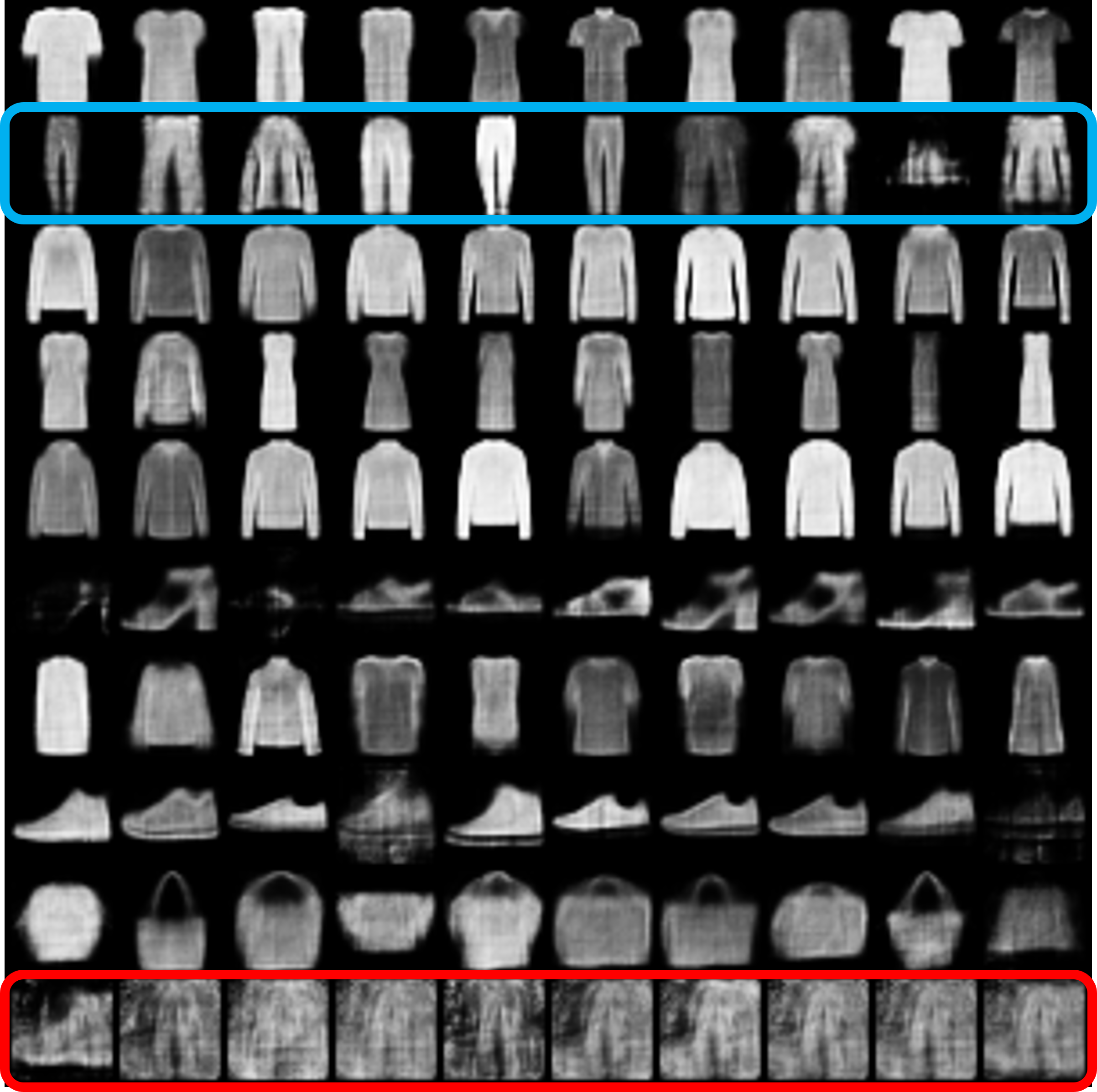}
    \caption{PM (white noise)+fine-tuning}
    \label{fig:sa_white_noise_ewc_fashion}
\end{subfigure}
\caption{The qualitative differences in images generated for all classes in the MNIST((a)-(d)) and Fashion-MNIST dataset ((e)-(h)) when the proposed method is applied to EWC and fine-tuning strategies. Specifically, the forgotten class \(c_{\rm f}\), denoted by the tenth class (MNIST: `1', Fashion-MNIST: `Ankle Boot'), is highlighted with a red border, and the newly learned \(c_{\rm new}\), denoted by the second class, is highlighted with a blue border. These classes are displayed in the second row from the top and the bottom row, respectively.}
\label{fig:sa_comparison_image}
\vspace{-7pt}
\end{figure*}

\section{Experiments}
\subsection{Settings}
In this experiment, we utilized the MNIST\footnote{http://yann.lecun.com/exdb/mnist/} and Fashion-MNIST\footnote{https://github.com/zalandoresearch/fashion-mnist?tab=readme-ov-file} datasets, to verify the effectiveness of our proposed method. These datasets are commonly used in the CIL task and are particularly employed in the demonstration of technical novelty.
MNIST and Fashion-MNIST consist of images from 10 classes, respectively. For the learning conditions in the GCIL setting, we set each model as follows: \(D_{\rm new}\) includes one specific class, \(D_{\rm f}\) includes another specific class different from \(D_{\rm new}\), and \(D_{\rm r}\) comprises the remaining eight classes.

The objective of this setting is to verify the effectiveness of our method across all 10 classes in multiple datasets. In this experiment, we validate the effectiveness by evaluating the performance of our method in all possible class configurations.

To evaluate the proposed framework, we used a simple one-hot Variational Autoencoder (VAE)~\cite{kingma2013auto} as the base deep generative model. 
Leveraging the simplicity and robustness of the one-hot VAE, we aim to demonstrate our approach's effectiveness in a comprehensive yet accessible manner.
We construct a model that inherently possesses the capability to retain knowledge about classes of \(c_{\rm f}\) and \(c_{\rm r}\). As we progress to the new class learning phase, the model is introduced to classes of \(c_{\rm r}\) and \(c_{\rm new}\), thus facilitating a comprehensive learning process. This phase consists of two principal approaches: the first involves strategically forgetting classes categorized as \(c_{\rm f}\), while the second emphasizes a straightforward learning method, devoid of any selective forgetting.

\setlength{\tabcolsep}{3.5pt} 
\begin{table*}[t]
\centering
\caption{The probability of correctly classifying each class in different methods by an external classifier, with fixed values of \(c_{\rm f}\) and \(c_{\rm new}\) in the MNIST dataset. PM (embedding) in the table refers to PM (embedding \(c_{\rm new}\)).}
\label{tab:mnist-model-accuracy-detailed}
\begin{tabular}{|l||c|c|c||c|c|c||c|c|c||c|c|c|}
\hline
Dataset : MNIST& \multicolumn{3}{c||}{(\(c_{\rm f}\),\(c_{\rm new}\))=(`9', `1')} & \multicolumn{3}{c||}{(\(c_{\rm f}\),\(c_{\rm new}\))=(`9', `2')} & \multicolumn{3}{c||}{(\(c_{\rm f}\),\(c_{\rm new}\))=(`8', `2')} & \multicolumn{3}{c|}{(\(c_{\rm f}\),\(c_{\rm new}\))=(`8', `1')} \\  
\hline
Model & \(c_{\rm f}(\downarrow)\) & \(c_{\rm r}(\uparrow)\) & \(c_{\rm new}(\uparrow)\) & \(c_{\rm f}(\downarrow)\) & \(c_{\rm r}(\uparrow)\) & \(c_{\rm new}(\uparrow)\) & \(c_{\rm f}(\downarrow)\) & \(c_{\rm r}(\uparrow)\) & \(c_{\rm new}(\uparrow)\) & \(c_{\rm f}(\downarrow)\) & \(c_{\rm r}(\uparrow)\) & \(c_{\rm new}(\uparrow)\) \\ \hline
Fine-tuning & 0.84& 0.90& 0.35& 0.82& 0.93& 0.46& 0.77& 0.93& 0.47& 0.78& 0.91& 0.35\\ \hline
Fine-tuning + PM (embedding)& 0.06& 0.91& \textbf{0.55}& 0.06& 0.96& \textbf{0.55}& 0.02& 0.93& \textbf{0.56}& 0.09& 0.91& \textbf{0.53}\\ \hline
Fine-tuning + PM (white noise)& 0.11& 0.91& \underline{0.38}& 0.14& 0.93& \underline{0.49}& 0.25& 0.93& \underline{0.49}& 0.15& 0.91& \underline{0.38}\\ \hline
EWC & 0.87& 0.90& 0.16& 0.86& 0.93& 0.27& 0.76& 0.93& 0.29& 0.81& 0.91& 0.16\\ \hline
EWC + PM (embedding)& 0.05& 0.91& \textbf{0.28}& 0.05& 0.92& \textbf{0.43}& 0.01& 0.92& \textbf{0.41}& 0.09& 0.91& \textbf{0.26}\\ \hline
EWC + PM (white noise)& 0.10& 0.91& \underline{0.19}& 0.10& 0.93& \underline{0.39}& 0.22& 0.93& \underline{0.39}& 0.14& 0.92& \underline{0.18}\\ \hline
\end{tabular}
\end{table*}

\begin{table*}[t]
\centering
\caption{The probability of correctly classifying each class in different methods by an external classifier, with fixed values of \(c_{\rm f}\) and \(c_{\rm new}\) in the Fashion-MNIST dataset. PM (embedding) in the table refers to PM (embedding \(c_{\rm new}\)).}
\label{tab:fashion-model-accuracy-detailed}
\begin{tabular}{|l||c|c|c||c|c|c||c|c|c||c|c|c|}
\hline
Dataset : Fashion-MNIST  & \multicolumn{3}{c||}{(\(c_{\rm f}\), \(c_{\rm new}\))=}& \multicolumn{3}{c||}{(\(c_{\rm f}\), \(c_{\rm new}\))=}& \multicolumn{3}{c||}{(\(c_{\rm f}\), \(c_{\rm new}\))=}& \multicolumn{3}{c|}{(\(c_{\rm f}\), \(c_{\rm new}\))=} \\
&\multicolumn{3}{c||}{(`Ankle Boot',`T-shirt/top')} & \multicolumn{3}{c||}{(`Ankle Boot',`Trouser')} & \multicolumn{3}{c||}{(`Bag',`Trouser')} & \multicolumn{3}{c|}{(`Bag', `T-shirt/top')} \\
\hline

Model & \(c_{\rm f}(\downarrow)\) & \(c_{\rm r}(\uparrow)\) & \(c_{\rm new}(\uparrow)\) & \(c_{\rm f}(\downarrow)\) & \(c_{\rm r}(\uparrow)\) & \(c_{\rm new}(\uparrow)\) & \(c_{\rm f}(\downarrow)\) & \(c_{\rm r}(\uparrow)\) & \(c_{\rm new}(\uparrow)\) & \(c_{\rm f}(\downarrow)\) & \(c_{\rm r}(\uparrow)\) & \(c_{\rm new}(\uparrow)\) \\ \hline
Fine-tuning & 0.94& 0.62& 0.48& 0.93& 0.66& 0.28& 0.82& 0.66& 0.28& 0.76& 0.63& 0.50\\ \hline
Fine-tuning + PM (embedding)& 0.00& 0.63& \textbf{0.56}& 0.27& 0.66& \textbf{0.55}& 0.13& 0.66& \textbf{0.40}& 0.01& 0.64& \textbf{0.55}\\ \hline
Fine-tuning + PM (white noise)& 0.17& 0.61& \underline{0.52}& 0.24& 0.66& \underline{0.29}& 0.32& 0.66& \underline{0.29}& 0.21& 0.63& \underline{0.53}\\ \hline
EWC & 0.92& 0.62& 0.17& 0.92& 0.66& 0.15& 0.84& 0.66& 0.15& 0.83& 0.63& 0.18\\ \hline
EWC + PM (embedding)& 0.00& 0.63& \textbf{0.28}& 0.25& 0.65& \textbf{0.43}& 0.12& 0.67& \textbf{0.41}& 0.02& 0.64& \textbf{0.27}\\ \hline
EWC + PM (white noise)& 0.08& 0.62& \underline{0.30}& 0.16& 0.66& \underline{0.19}& 0.27& 0.70& \underline{0.19}& 0.20& 0.64& \underline{0.31}\\ \hline
\end{tabular}
\end{table*}

In addition, we incorporate an additional layer of complexity in our experiment by testing two variations of the forgetting process. In one variation, forgetting is simulated by introducing a white noise image into the data associated with \(D_{\rm f}\). 
For the second variation, we incorporate images from the newly learned class \(c_{\rm new}\) into \(c_{\rm f}\) to analyze the impact of class confusion.
The dual variation aims to further investigate the effectiveness of selective forgetting. In particular, we focus on how different modes of forgetting influence the model's learning efficiency in a newly learned class and the impact on other existing classes.

For a comprehensive evaluation, we employ an external classifier to analyze the samples generated from the trained one-hot VAEs for each class. This analysis aims to confirm the extent to which generated images retain information pertinent to specific classes. Specifically, the metric for evaluating the accurate classification of a class is given by

\begin{equation}
    E_{p(x|\theta,c_{\text{class}})} [P_{\phi}(y = c_{\text{class}} | x)],
\end{equation}
where this expectation is over the samples generated by our trained model, and $P_{\phi}(y|x)$ represents the probability attributed by a pre-trained external classifier. Assuming $q(x|c_{\rm {\text{class}}})$ to be a non-informative distribution such as a uniform distribution, this probability should theoretically converge to $\frac{1}{N_{\text{class}}}$(\(=\frac{1}{10}\)) for the datasets.

It is important to note that the classification criteria differ significantly between the classes designated for forgetting (\(c_{\rm f}\)) and other classes (\(c_{\rm r}\), \(c_{\rm new}\)). The evaluation for each class is conducted separately, with an aggregation (mean calculation) applied across the multiple \(c_{\rm r}\). For \(c_{\rm f}\), the desired expectation is close to or smaller than \(\frac{1}{10}\) because of aims to forget the original information. Conversely, for \(c_{\rm r}\) and \(c_{\rm new}\), the desired outcome is close to 1 more desirable because of aims to either retain pre-existing knowledge or acquire new information more accurately, respectively. 


This study also investigates the adaptability of pre-existing models to learn new classes through the application of fine-tuning procedures. Fine-tuning serves as a baseline method in CIL. It has been reported that fine-tuning can achieve better performance than the existing CIL method in accuracy in specific tasks~\cite{masana2022class}.
During the fine-tuning process, the model trains without class \(c_{\rm f}\). This approach is designed to isolate the effects of fine-tuning on the model's performance, allowing us to conclude its impact on the model's ability to incorporate new knowledge.

\subsection{Results}


We show the impacts of the performance of newly learned classes \(c_{\rm new}\) with a forgetting mechanism and GCIL in Fig.~\ref{fig:conf_matrix_comparison}.  
In Fig.\ref{fig:conf_matrix_comparison}, the vertical axis represents the classes that were forgotten (\(c_{\rm f}\)), while the horizontal axis denotes the classes being learned (\(c_{\rm new}\)). 
Each figure shows the relationship between the forgotten class \(c_{\rm f}\) with white noise image and learned class \(c_{\rm new}\).
As shown in Fig. 2, the introduction of the forgetting mechanism led to an improvement in the accuracy of \(c_{rm new}\) under most conditions.
The results indicate a notable enhancement in accuracy for both MNIST and Fashion-MNIST datasets within the incorporation of the PM. The performance improvement is observed across both EWC and fine-tuning learning strategies. These results show a general benefit of integrating the PM. In particular, the application of the PM before the implementation of EWC shows a more apparent accuracy improvement compared to its application in fine-tuning. Furthermore, the extent of accuracy enhancement is observed to be dependent on the specific newly learned class \(c_{\rm new}\). This is underscoring the influence of class characteristics on the effectiveness of the PM. 

Next, the qualitative comparison between EWC, fine-tuning, and PM (white noise) is presented in Fig.~\ref{fig:sa_comparison_image}. These figures show the output for each class when trained using these respective methods.
Figures~\ref{fig:sa_comparison_image} (a)-(d) show PM (embedding \(c_{\rm new}\)) from MNIST dataset successfully integrates \(c_{\rm new}\) the number `1' into tenth class, effectively preserving the information of the retained classes \(c_{\rm new}\).
This integration not only demonstrates the preservation of existing class information but also confirms the effectiveness of GCIL.
In the case of the Fashion-MNIST dataset, Figs.~\ref{fig:sa_comparison_image} (e) - (h) show that PM (embedding \(c_{\rm new}\)) substitutes the \(c_{\rm new}\) `Ankle Boot' with an image marked by the white noise image. Furthermore, the class `Trouser' identified as second class, is visible in the second row from the top and produced by both the comparative methods and the application of PM (embedding \(c_{\rm new}\)).
This observation shows that the PM's applicability is irrespective of the specific dataset or the data being embedded in diverse GCIL scenarios.

The quantitative evaluation is shown in Table \ref{tab:mnist-model-accuracy-detailed} and Table \ref{tab:fashion-model-accuracy-detailed}. From Table \ref{tab:mnist-model-accuracy-detailed}, we can see that the application of the PM (embedding \(c_{\rm new}\)) demonstrates superior learning accuracy for \(c_{\rm new}\) compared to conventional methods, as indicated by the bold probability of class \(c_{\rm new}\). 
This enhanced learning shows the efficacy of embedding \(c_{\rm new}\) in learning new class information. Furthermore, the observation of \(D_{\rm f}\) values being lower than \(1/10\) suggests that the PM replaces the information of \(c_{\rm f}\) with other classes and it is effective in erasing outdated information.

However, it is noted that in generative models, which demand higher performance in information representation, there are limitations in embedding semantically meaningful information. In such cases, as the underlined probability of class \(c_{\rm new}\), the other choice of non-semantic white noise image still results in better learning accuracy for \(c_{\rm new}\) than existing methods. This observation indicates that PM maintains its effectiveness in learning new classes, even with the introduction of non-semantic images that are not directly related to the other classes.

As detailed in Table \ref{tab:fashion-model-accuracy-detailed}, the difference in accuracy achieved by integrating the proposed method with existing techniques is shown in the Fashion-MNIST dataset. 
This shows the effectiveness of the proposed method in multiple datasets.
It is important to note that this difference in performance improvement when the different class \(c_{\rm new}\) such as `Trouser' and `T-shirt/top' are newly learned. This difference in performance improvement indicates that the performance improvement of the PM depends on the specific characteristics of the class and the complexities of the dataset.
In summary, introducing a forgetting mechanism to existing learning methods improves the performance of the existing method in most cases.

\section{Conclusion} 
In this paper, we have presented a novel approach to GCIL by introducing a forgetting mechanism, significantly enhancing model performance. The proposed method promises to improve GCIL by strategically discarding outdated or irrelevant information, offering a novel way to balance between acquiring new knowledge and preserving essential past information. 
Through experiments with various datasets, we have demonstrated the effectiveness of incorporating a forgetting mechanism within the context of the GCIL task. The proposed method is anticipated to serve as a key technique, particularly beneficial in scenarios of constrained model capacity or during the process of continuous knowledge acquisition.

\vfill\pagebreak
\clearpage

\bibliographystyle{IEEEbib}
\bibliography{refs}

\end{document}